\documentclass[letterpaper, 10 pt, conference]{ieeeconf}  

\IEEEoverridecommandlockouts                              

\usepackage{graphicx}
\usepackage{amsmath,amssymb}
\usepackage{algorithm}
\usepackage{algorithmicx}
\usepackage{algpseudocode}
\usepackage{url}
\usepackage{subcaption}
\usepackage{enumerate}

\title{\LARGE \bf
Robustifying Model-Based Locomotion by Zero-order Stochastic Nonlinear Model Predictive Control with Guard Saltation Matrix
}

\author{Sotaro Katayama$^{1}$, Noriaki Takasugi$^{1}$, Mitsuhisa Kaneko$^{2}$, Norio Nagatsuka$^{3}$, and Masaya Kinoshita$^{1}$
\thanks{$^{1}$Sony Group Corporation, Minato-ku, Tokyo, Japan, 108-0075
        {\tt\small sotaro.katayama@sony.com}}%
\thanks{$^{2}$Sony Global Manufacturing and Operations Corporation, Minato-ku, Tokyo, Japan, 108-0075}
\thanks{$^{3}$Sony Interactive Entertainment Inc., Sony City 1-7-1, Konan, Minato-ku, Tokyo, 108-0075 Japan}
}

\begin{document}

\maketitle
\thispagestyle{empty}
\pagestyle{empty}

\begin{abstract}

This paper presents a stochastic/robust nonlinear model predictive control (NMPC) to enhance the robustness of model-based legged locomotion against contact uncertainties.
We integrate the contact uncertainties into the covariance propagation of stochastic/robust NMPC framework by leveraging the guard saltation matrix and an extended Kalman filter-like covariance update.
We achieve fast stochastic/robust NMPC computation by utilizing the zero-order algorithm with additional improvements in computational efficiency concerning the feedback gains.
We conducted numerical experiments and demonstrate that the proposed method can accurately forecast future state covariance and generate trajectories that satisfies constraints even in the presence of the contact uncertainties.
Hardware experiments on the perceptive locomotion of a wheeled-legged robot were also carried out, validating the feasibility of the proposed method in a real-world system with limited on-board computation.

\end{abstract}

\section{INTRODUCTION}
A fundamental challenge in model-based legged locomotion is ensuring robustness against uncertainties that arise in contacts with the environment.
When traversing uneven terrains, robots must determine actions based on information acquired from exteroceptive sensors such as RGB-D cameras or LiDAR.
However, the accuracy of this environmental information, such as the terrain height surrounding the robot, is imperfect due to limitations of sensor resolution, occlusions, or non-flat terrains.
Model-based control methods, such as model predictive control (MPC), have struggled with estimation inaccuracies in environmental information.
For instance, if the estimated terrain height is higher than the actual height, a swinging foot may fail to establish contact, leading to the robot's instability and potential falls.

In contrast to model-based control methods, model-free reinforcement learning (RL) offers the ability to incorporate various environmental uncertainties into control policies.
In a study by Miki et al. \cite{miki2022learning}, the authors propose explicitly encoding uncertainties in exteroceptive sensors, such as the terrain height map around the robot, into the training process.
By employing domain randomization to account for perception uncertainties and environmental variations, the policy achieves the capability to traverse challenging real-world environments.
This RL approach to mitigating uncertainties prompts a question: How can model-based control methods achieve robustness against contact uncertainties?

In this paper, we explore a model-based control approach that explicitly considers contact uncertainties in legged locomotion.
Specifically, we incorporate the contact uncertainties into stochastic/robust nonlinear MPC (NMPC) by leveraging the guard saltation matrix introduced by Payne et al. \cite{payne2022uncertainty} and the extended Kalman filter (EKF)-like covariance propagation proposed by Zhang and Ohtsuka \cite{zhang2023output}.
We then apply the zero-order stochastic/robust NMPC algorithm \cite{zanelli2021zero,messerer2021efficient,zhang2023output} with additional enhancements to computational efficiency regarding the feedback gains.
Our stochastic/robust NMPC ensures robustness to contact uncertainty via online optimization, guaranteeing the satisfaction of constraints such as the friction cone, collision avoidance, and joint limits even in the presence of disturbances in the contact events.

\begin{figure}[t]
    \centering
    \includegraphics[scale=0.4]{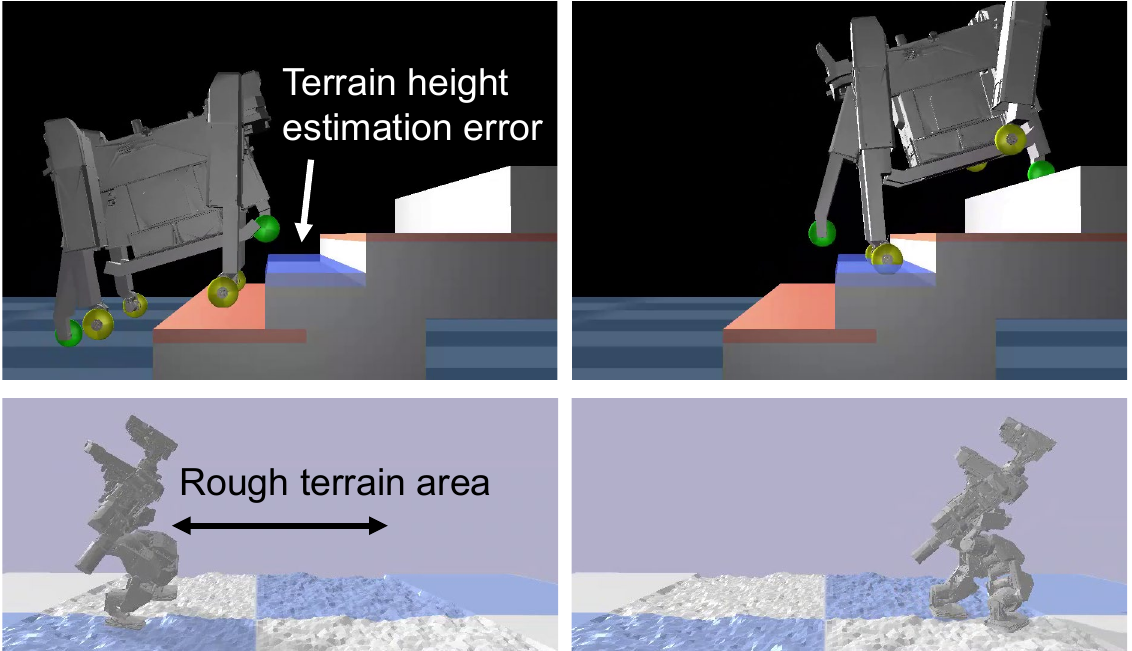}
    \caption{Simulations utilizing the proposed stochastic nonlinear model predictive control.
    Top row: Tachyon 3, a six-telescopic-wheeled-legged robot, ascends stairs under terrain height estimation errors.
    Bottom row: EVAL-03, a miniature humanoid robot, traverses a rough terrain area.}
  \label{fig:terrainHeightEstimationError}
\end{figure}

\subsection{Related Works}
Model-based control methods for legged robots, predominantly MPC, inevitably involve uncertainties.
Even besides the contact-related ones, there are a variety of uncertainties such as model approximations (e.g., single rigid-body approximation), model parameters (e.g., inertias), and unmodeled effects (e.g., actuator dynamics).
Stochastic/robust MPC is a promising framework to cope with these uncertainties \cite{mesbah2016stochastic}.
Stochastic/robust MPC predicts the uncertainties in the system state over the horizon and tightens constraints to ensure that the constraints are satisfied even in the presence of disturbances.
Despite the potential computational burden, the advent of zero-order algorithms has paved the way for practical applications \cite{zanelli2021zero,messerer2021efficient,zhang2023output}.

In \cite{xu2023robust,gazar2023multi}, the stochastic/robust MPC techniques have been successfully applied for legged locomotion.
Xu et al. \cite{xu2023robust} employ linear robust MPC to account for uncertainties arising from single rigid-body and linear approximations of the system dynamics.
Gazar et al. \cite{gazar2023multi} also utilize stochastic NMPC to enhance robustness against uncertainties in the continuous-time system dynamics.
However, these studies cannot capture the uncertainties involved in the contact events.
Incorporating uncertainties related to contacts into stochastic NMPC is not straightforward compared to simple process noise in continuous-time system dynamics.

The saltation matrix, a mathematical tool to analyze the sensitivity of the state with respect to discrete events, can be a promising tool to capture the uncertainties in the contact events \cite{kong2023hybrid}.
It has been utilized in offline trajectory optimization \cite{tucker2023robust}, (deterministic) contact-implicit MPC \cite{kong2023hybrid}, and state estimation \cite{payne2022uncertainty}.
Notably, Payne et al. \cite{payne2022uncertainty} propose the guard saltation matrix, an extension of the original saltation matrix that can incorporate uncertainty in the guard condition, such as terrain height in the case of legged locomotion.

\subsection{Contribution}
In this paper, we present a stochastic/robust NMPC against contact uncertainties for legged locomotion.
Our contributions can be summarized as follows:
\begin{itemize}
    \item A novel covariance propagation for stochastic/robust NMPC against contact uncertainties leveraging a priori covariance update using the guard saltation matrix and EKF-like a posteriori covariance update.
    \item An efficient numerical algorithm for stochastic/robust NMPC utilizing the zero-order algorithm and interior point method.
    \item Application of the proposed method to the full-centroidal NMPC formulation.
    \item Performance comparison between different NMPC methods through simulations on two robotic platforms.
    \item Hardware experiments on perceptive locomotion over uneven terrains with on-board implementation of the proposed stochastic/robust NMPC.
\end{itemize}

This paper is organized as follows.
Section \ref{sec:saltation} briefly summarizes the saltation matrices for hybrid systems.
Section \ref{sec:smpc} proposes a stochastic/robust NMPC framework leveraging the saltation matrices and EKF-like covariance update for contact uncertainties.
Section \ref{sec:algo} introduces an efficient computation strategy to implement the proposed stochastic/robust NMPC.
Section \ref{sec:experiments} investigates the effectiveness of the proposed approach through simulations and hardware experiments.
Finally, Section \ref{sec:conclusions} concludes this paper and discusses future works.

\section{HYBRID SYSTEM AND GUARD SALTATION MATRIX}\label{sec:saltation}
In this section, we summarize the guard saltation matrix introduced in \cite{payne2022uncertainty} that characterizes a certain aspect of legged locomotion.
To begin, we introduce a mode index $m$ to represent the combination of contact flags (on/off) for all feet.
For the state $x (t) \in \mathbb{R}^{n_x}$ and control input $u (t) \in \mathbb{R}^{n_u}$, the continuous-time state equation, or \textit{flow map}, for mode $m$ is defined as follows:
\begin{equation}\label{eq:flow}
        \dot{x} (t) = f_m (t, x(t), u(t)).
\end{equation}
The switch from mode $m$ to mode $m+1$ occurs when the system satisfies the \textit{guard condition}:
\begin{equation}\label{eq:guard}
    g_{(m, m+1)} (t_{(m, m+1)} ^-, x(t_{(m, m+1)} ^-)) = 0,
\end{equation}
where $t_{(m, m+1)} ^-$ denotes the instant immediately before the switch from mode $m$ to mode $m+1$.
At the instance of the switch, the state undergoes an instantaneous change, or a state jump, represented by the \textit{reset map}:
\begin{equation}\label{eq:reset}
    x(t_{(m, m+1)} ^+) = R_{(m, m+1)}(t_{(m, m+1)}^-, x(t_{(m, m+1)}^-)),
\end{equation}
where $t_{(m, m+1)}^+$ is the instant immediately after the switch.
The guard condition (\ref{eq:guard}) represents, for example, the distance between a foot and the ground, while the reset map (\ref{eq:reset}) can capture impulsive dynamics due to collisions \cite{kong2023saltation}.

The original saltation matrix \cite{kong2023saltation} describes the first-order sensitivity of $x(t_{(m, m+1)} ^+)$ with respect to $x(t_{(m, m+1)}^-)$ in cases where (\ref{eq:guard}) is one-dimensional, i.e., (\ref{eq:guard}) is a scalar-valued function. Specifically, the saltation matrix for the switch from mode $m$ to $m+1$ is defined as \cite{kong2023saltation}:
\begin{align}\label{eq:saltation}
    & \Xi_{x, (m, m+1)} := \nabla_x R_{(m, m+1)} \notag \\ 
    & + \frac{ f_{m+1} ^+ - \nabla_x R_{(m, m+1)} f_m ^{-} - \nabla_t R_{(m, m+1)} }{\nabla_t g_{(m, m+1)} + \nabla_x g_{(m, m+1)} f_m ^{-}} \nabla_x g_{(m, m+1)},
\end{align}
where $f_m^- := f_m(t_{(m, m+1)}^-, x(t_{(m, m+1)}^-), u(t_{(m, m+1)}^-))$ and $f_{m+1}^+ := f_{m+1}(t_{(m, m+1)}^+, x(t_{(m, m+1)}^+), u(t_{(m, m+1)}^+))$, and we drop the obvious arguments from $g_{(m, m+1)}$ and $R_{(m, m+1)}$.
Similarly, the guard saltation matrix \cite{payne2022uncertainty} describes the first-order sensitivity of $x(t_{(m, m+1)}^+)$ with respect to the guard condition and is defined as:
\begin{equation}\label{eq:guardSaltation}
    \Xi_{g, (m, m+1)} := 
            \frac{\nabla_x R_{(m, m+1)} f_m ^{-} + \nabla_t R_{(m, m+1)} - f_{m+1} ^+}{\nabla_t g_{(m, m+1)} + \nabla_x g_{(m, m+1)} f_m ^{-}}.
\end{equation}
Note that, to define these saltation matrices, the following transversality condition must be hold: 
\begin{equation}\label{eq:transversality}
    \nabla_t g_{(m, m+1)} + \nabla_x g_{(m, m+1)} f_m ^{-} < 0.
\end{equation}

\section{STOCHASTIC/ROBUST MODEL PREDICTIVE CONTROL WITH SALTATION MATRIX}\label{sec:smpc}

\subsection{Problem Settings}

\begin{figure}[tb]
    \centering
    \includegraphics[scale=0.75]{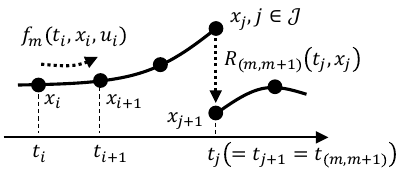}
    \caption{Discretization of the optimal control problem.}
  \label{fig:discretization}
\end{figure}

We formulate the optimal control problem (OCP) for a legged robot, assuming a predetermined contact sequence composed of a mode sequence $1, 2, ..., M$ and corresponding switching times $t_{(1, 2)}, ..., t_{(M-1, M)}$.
The finite-time horizon is discretized into a $N + 1$ ($N > 0$) nodes whose instants are given by $[t_0, t_1, ..., t_N]$.
We define $\mathcal{J}$ as the set of jump nodes $j$ where a state jump occurs between $j$ and $j + 1$, i.e., $t_{j} = t_{j+1} = t_{(m, m+1)}$, with $m$ as the corresponding mode index.
Additionally, we define $\mathcal{I} := \left\{ 0, 1, ..., N-1 \right\} \setminus \mathcal{J}$.
Employing the direct multiple-shooting method \cite{bock1984multiple}, we introduce state trajectory $x_0, ..., x_N \in \mathbb{R}^{n_x}$ and control inputs trajectory $u_i \in \mathbb{R}^{n_u}$ for all $i \in \mathcal{I}$.
The definition of the variables in our discretization are summarized in Fig. \ref{fig:discretization}.

In stochastic/robust settings, we account for process noise in the discrete-time state equation.
With a process noise trajectory $w_0, ..., w_{N-1} \in \mathbb{R}^{n_w}$, the continuous-time state equation (\ref{eq:flow}) is discretized into the following discrete-time state equation:
\begin{equation}\label{eq:stateEquation}
    x_{i+1} = F_i (t_i, x_i, u_i, w_i), \; i \in \mathcal{I}.
\end{equation}
For instance, using the forward Euler method, the discrete-time state equation is:
\begin{equation*}
    F_i (t_i, x_i, u_i, w_i) := x_i + f_m (t_i, x_i, u_i) \Delta \tau_i + \Gamma_i w_i , \; i \in \mathcal{I},
\end{equation*}
where $\Gamma_i \in \mathbb{R}^{n_x \times n_w}$ is a constant matrix, $m$ is the appropriate mode index, and $\Delta \tau > 0$ is the time step.
We also consider disturbances in the state jump:
\begin{equation}\label{eq:discretStateEquation}
    x_{j + 1} = F_j (t_j, x_j, w_j) := R_{(m, m+1)} (t_j, x_j) + \Gamma_j w_j, \; j \in \mathcal{J}.
\end{equation}

Following the literature on stochastic/robust NMPC \cite{mesbah2016stochastic,zanelli2021zero,messerer2021efficient}, we assume that the state is represented by a Gaussian distribution.
We introduce the mean and covariance of the state trajectory $\bar{x}_0, ..., \bar{x}_N \in \mathbb{R}^{n_x}$ and $P_0, ..., P_N \in \mathbb{R}^{n_x \times n_x}$.
The covariance propagation is approximated as:
\begin{align}\label{eq:covarianceEquation}
    P_{i+1} &= (A_i + B_i K_i) P_i (A_i + B_i K_i)^{\rm T} + \Gamma_i W_i \Gamma_i ^{\rm T} \notag \\ 
            &:= \Phi_i (P_i, K_i, \bar{x}_i, u_i), \;\; i \in \mathcal{I},
\end{align}
where $A_i := \nabla_x F_i (t_i, \bar{x}_i, u_i, 0)$, $B_i := \nabla_u F_i (t_i, \bar{x}_i, u_i, 0)$, $\Gamma_i := \nabla_w F_i (t_i, \bar{x}_i, u_i, 0)$, $W_i \in \mathbb{R}^{n_w \times n_w}$ is a symmetric positive semi-definite matrix, and $K_i \in \mathbb{R}^{n_u \times n_x}$ is a state feedback gain.
Equation (\ref{eq:covarianceEquation}) refers to closed-loop covariance propagation with an appropriate feedback gain $K_i$, and reduces to open-loop propagation with $K_i = O$.
Closed-loop propagation can mitigate the overgrowth of state covariance and result in less conservative constraints \cite{messerer2021efficient}, but requires optimization of the feedback gain $K_i$.
In robust MPC, $W_i$ describes an ellipsoid containing the process noise $w_i$ \cite{zanelli2021zero, messerer2021efficient}.
In stochastic MPC, $W_i$ describes the noise distribution covariance, such as $w_i \sim \mathcal{N} (0, W_i)$ \cite{mesbah2016stochastic}.

\subsection{Covariance Propagation Between a Contact Event}
Covariance propagation between mode switches (contact events) is crucial for handling contact uncertainties.
We propose a novel covariance update consisting of an \textit{a priori} update using saltation matrices and an \textit{a posteriori} update akin to the EKF.

\subsubsection{A priori covariance update using Guard Saltation Matrices}
We utilize saltation matrices (\ref{eq:saltation}) and (\ref{eq:guardSaltation}) in covariance propagation to introduce contact uncertainties.
When the guard condition (\ref{eq:guard}) is one-dimensional, as in EKF applications \cite{payne2022uncertainty}, we update the covariance by
\begin{align}\label{eq:singleJumpCovarianceEquation}
    \hat{P}_{j+1}
    = \;\, & \Xi_{x, (m, m+1)} \; P_j \; {\Xi_{x, (m, m+1)}} ^{\rm T} \notag \\
    & \;\; + \; \Xi_{g, (m, m+1)} \; C_{g} \; {\Xi_{g, (m, m+1)}} ^{\rm T},
\end{align}
where $C_{g} \in \mathbb{R}$ denotes the covariance of $g_{(m, m+1)}$.
However, typical gaits such as quadrupedal trotting, several feet make contact simultaneously at the same instant, which requires multi-dimensional guard conditions.

Unfortunately, literature on the saltation matrices such as \cite{payne2022uncertainty} only treat the cases where the guard condition (\ref{eq:guard}) is one-dimensional.
To apply the saltation matrix even with the multi-dimensional guard conditions whose dimension is $n_g > 0$, we update the covariance as:
\begin{align}\label{eq:multipleJumpCovarianceEquation}
    \hat{P}_{j+1}
    = \sum_{i = 1}^{n_g} \; & \left\{ \Xi_{x, (m, m+1)}^{(i)} \; P_j \; {\Xi_{x, (m, m+1)} ^{(i)}} ^{\rm T} \right. \notag \\
    & \;\; \left. + \; \Xi_{g, (m, m+1)} ^{(i)} \; C_{g ^{(i)}} \; {\Xi_{g, (m, m+1)} ^{(i)}} ^{\rm T} \right\},
\end{align}
where $\Xi_{x, (m, m+1)}^{(i)}$ and $\Xi_{g, (m, m+1)}^{(i)}$ are the saltation matrices for the $i$-th element of $g_{(m, m+1)}$, and $C_{g^{(i)}} \in \mathbb{R}$ is the covariance of the $i$-th element of the guard condition $g_{(m, m+1)}$.
This update represents a summation of the one-dimensional update (\ref{eq:singleJumpCovarianceEquation}) for each element of the guard condition.
In practice, $g_{(m, m+1)} ^{(i)}$ does not necessarily satisfy the transversality condition (\ref{eq:transversality}). 
We ommit such index $i$ from the summation of (\ref{eq:multipleJumpCovarianceEquation}).
While this approach was effective in our experiments, theoretically-supported methods such as using the Bouligand derivative \cite{burden2016event} are left for future work.

Note that computing (\ref{eq:saltation}) and (\ref{eq:guardSaltation}) requires values of $f_m ^{-}$ and $f_{m+1} ^{+}$.
We approximate them by using variables on the nearest discretization nodes, i.e., by $f_m (t_{j-1}, x_{j-1}, u_{j-1})$ and $f_{m+1} (t_{j+1}, x_{j+1}, u_{j+1})$, respectively.

\subsubsection{A posteriori covariance update using output-feedback MPC}
The update (\ref{eq:multipleJumpCovarianceEquation}) is an open-loop covariance update and may lead to overgrowth of the covariance.
To address this, we apply an a posteriori covariance update from output-feedback stochastic MPC \cite{zhang2023output}.
For the stochastic process whose belief dynamics are (\ref{eq:stateEquation}) and (\ref{eq:discretStateEquation}), the guard condition (\ref{eq:guard}), i.e., the contact positions, can be regarded an output equation, as is well-known in the EKF-based state estimation for the legged robots \cite{bloesch2013state}.
Treating the guard condition (\ref{eq:guard}) as an output equation, we apply the a posteriori covariance update law proposed in \cite{zhang2023output}, resulting in:
\begin{align}\label{eq:outputCovarianceEquation_L}
    L_{j+1}
    := \; & \hat{P}_{j+1} \nabla_x g_{(m, m +1)} ^{\rm T} \cdot \notag \\
     & (\nabla_x g_{(m, m +1)} \hat{P}_{j+1} \nabla_x g_{(m, m +1)} ^{\rm T} + C_g) ^{-1},
\end{align}
\begin{equation}\label{eq:outputCovarianceEquation_P}
    {P}_{j+1} = (I - L_{j+1} \nabla_x g_{(m, m +1)}) \hat{P}_{j+1},
\end{equation}
where $\hat{P}$ is the covariance after the update by (\ref{eq:multipleJumpCovarianceEquation}).
We summarize the covariance update from $P_j$ to $P_{j+1}$ using (\ref{eq:multipleJumpCovarianceEquation}), (\ref{eq:outputCovarianceEquation_L}), and (\ref{eq:outputCovarianceEquation_P}) into ${P}_{j+1} = \Phi_j (P_j, \bar{x}_j)$.
This a posteriori update can also be applied for $i \in \mathcal{I}$ to suppress the covariance overgrowth, although this paper focuses on the update between a contact event.

\subsection{Stochastic/Robust Optimal Control Problem}
With the settings described, the stochastic/robust OCP is formulated as:
\begin{subequations}\label{eq:ocp}
\begin{align}
    \min_{\substack{\bar{x}_0, ..., \bar{x}_N, \\ 
                    \bar{u}_0, ..., \bar{u}_{N-1}, \\ 
                    P_0, ..., P_{N}}} 
    & J := V_N (\bar{x}_N) + \sum_{i \in I} l_i (\bar{x}_i, \bar{u}_i) + \sum_{j \in J} l_j (\bar{x}_i) \\
    {\rm s.t.} \;\; & \bar{x}_{0} = \bar{x}, \label{subeq:x0} \\ 
    & \bar{x}_{i+1} = F_i (t_i, \bar{x}_i, \bar{u}_i, 0), \;\; i \in I, \label{subeq:Fi} \\ 
    & \bar{x}_{j+1} = F_j (t_j, \bar{x}_j, 0), \;\; j \in J, \label{subeq:Fj} \\ 
    & P_{i+1} = \Phi_i (P_i, K_i, \bar{x}_i, \bar{u}_i), \;\; i \in I, \label{subeq:Pi} \\ 
    & P_{j+1} = \Phi_j (P_j, \bar{x}_j), \;\; j \in J, \label{subeq:Pj} \\ 
    & g_j (\bar{x}_j) = 0 , \;\; j \in J, \label{subeq:g} \\ 
    & h_i (\bar{x}_i, \bar{u}_i) + \beta_i (P_i, K_i, \bar{x}_i, \bar{u}_i) \geq 0 , \; i \in I, \label{subeq:hi} \\ 
    & h_j (\bar{x}_j) + \beta_j (P_j, \bar{x}_j) \geq 0 , \; j \in J, j = N, \label{subeq:hj} 
\end{align}
\end{subequations}
where $V_{N} (\cdot)$ denotes the terminal cost, $l_{i} (\cdot)$ and $l_{j} (\cdot)$ denote the stage costs, and $h_i (\cdot)$ and $h_j (\cdot)$ denote the inequality constraint, respectively.
$\beta_i (\cdot)$ and $\beta_j (\cdot)$ are backoff terms introduced to ensure that the original inequality constraints $h_i (\cdot) \geq 0$ and $h_j (\cdot) \geq 0$ are satisfied under the predicted state covariances $P_i$ and $P_j$, respectively, and are defined as:
\begin{align}\label{eq:backoff}
    \beta_i & (P_i, K_i, \bar{x}_i, \bar{u}_i) := \notag \\ 
    & \gamma_i
    \sqrt{
        \nabla_{x, u} h_i (x_i, u_i) ^{\rm T} 
        \begin{bmatrix}
            I \\ K_i
        \end{bmatrix}
        P_i
        \begin{bmatrix}
            I \\ K_i
        \end{bmatrix} ^{\rm T}
        \nabla_{x, u} h_i (x_i, u_i) 
    },
\end{align}
\begin{equation}\label{eq:backoffJump}
    \beta_j (P_j, \bar{x}_j) :=
    \gamma_j
    \sqrt{
        \nabla_{x} h_j (x_j) ^{\rm T} 
        P_j
        \nabla_{x} h_j (x_j) 
    }.
\end{equation}
In robust MPC, $\gamma_i$ is typically set to 1.0.
In stochastic MPC, for a chance constraint:
\begin{equation}\label{eq:chanceConstraint}
    {\rm Pr} [h_i (x_i, u_i) \geq 0] > p_i,
\end{equation}
where, $1.0 > p_i > 0.5$ is the probability of constraint satisfaction, the constraint is approximated as a deterministic constraint with $\gamma_i := \Psi^{-1} (p_i)$, where $\Psi^{-1} (\cdot)$ is the inverse cumulative density function of a standard normal Gaussian distribution \cite{zhang2023output}.
Note that $\Psi^{-1} (\cdot)$ can be computed numerically offline.

\section{EFFICIENT STOCHASTIC/ROBUST MODEL PREDICTIVE CONTROL ALGORITHM}\label{sec:algo}

\subsection{Zero-Order Algorithm for Covariance Propagation}
The zero-order algorithm of stochastic/robust NMPC \cite{zanelli2021zero, messerer2021efficient,zhang2023output} simplifies the Newton-type iteration for the OCP (\ref{eq:ocp}) by disregarding the sensitivities of the covariance propagation equations (\ref{subeq:Pi}) and (\ref{subeq:Pj}) and the sensitivities of the backoff terms $\beta (\cdot)$.
The computational burden per Newton-type iteration is competitive with the nominal NMPC.

A limitation of the zero-order algorithm is that its convergence rate may slow down when the process noise covariance $W_i$ is substantial, as the backoff terms have a large effect on the convergence.
We empirically found that for efficient optimization, $W_i$ should be carefully tuned to moderate the covariance propagation rather than accurately reflecting the process noise covariance of the actual system.

\subsection{Interior Point Method and Feedback Gain}
We employ the nonlinear primal-dual interior point method to handle hard inequality constraints \cite{nocedal1999numerical}.
It should be noted that state-only inequality constraints need to be treated as soft constraints such as \cite{feller2016relaxed} to prevent the backoff term (\ref{eq:backoff}) from becoming excessively large, which leads to no feasible solution satisfying the tightened constraints (\ref{subeq:hi})–(\ref{subeq:hj}) \cite{zanelli2021zero}.
In practice, we also might have to clip the maximum values of the backoff terms to avoid ill-conditioned optimization problems.

The Riccati recursion \cite{rawlingsmodel}, used in conjunction with the interior point method and soft constraints, allows for efficient computation of the Newton step and provides the feedback gain trajectory $K_0, ..., K_{N-1}$ as a by-product.
This method facilitates the efficient online update of the feedback gain during MPC application, particularly without additional computations than the Newton-step computation solely to retrieve the state feedback gains or backoff terms, which are required in the previous studies \cite{messerer2021efficient, xu2023robust}.

In applying the proposed method in MPC, the optimal feedback gain trajectory $K_0, ..., K_{N-1}$ is stored after each MPC iteration.
At the next time step, the feedback gain trajectory is estimated (i.e., warmstarted) using the stored feedback gain trajectory as well as the state and control input trajectories, for example, through linear interpolation.

\subsection{Algorithm Summary}

\begin{figure}[t]
  \begin{algorithm}[H]
    \caption{Zero-Order Stochastic/Robust NMPC}
    \label{algo}
    \begin{algorithmic}[1]
    \Require{Current time $t$ and current state $x (t)$}
    \Ensure{Optimal trajectories $x_0, ..., x_{N}$ and $u_0, ..., u_{N-1}$}
    \State Initialize $x_0, ..., x_{N}$, $u_0, ..., u_{N-1}$, and $K_0, ..., K_{N-1}$ by the linear interpolation of the previous NMPC solution.
    \While{termination criteria are not satisfied}
    \State Compute the values and partial derivatives of $V_N (\cdot)$, $l_i (\cdot)$, $l_j (\cdot)$, $F_i (\cdot)$, $F_j (\cdot)$, $g_j (\cdot)$, $h_i (\cdot)$, and $h_j (\cdot)$.
    \State Propagate the state covariances by (\ref{subeq:Pi}) and (\ref{subeq:Pj}).
    \State Compute backoff terms $\beta_i (\cdot)$ by (\ref{eq:backoff}) and (\ref{eq:backoffJump}).
    \State Reduce the inequality constraints into the cost terms, e.g., via nonlinear interior point method.
    \State Perform the Riccati recursion to compute the search directions $\Delta x_0, \allowbreak ..., \allowbreak \Delta x_{N}$, $\Delta u_0, \allowbreak ..., \allowbreak \Delta u_{N-1}$ and feedback gains $K_0, \allowbreak ..., \allowbreak K_{N-1}$.
    \State Find a step-size via line-search (optional).
    \State Update and store $x_0, ..., x_{N}$ and $u_0, ..., u_{N-1}$.
    \State Update the barrier parameter (optional).
    \EndWhile
    \end{algorithmic}
  \end{algorithm}
\end{figure}

Algorithm \ref{algo} outlines the proposed zero-order stochastic/robust NMPC with the interior point method.
The algorithm begins by initializing (warmstarting) state, control input, and feedback gain trajectories using linear interpolation of the previous NMPC solution (line 1).
It then iterates (lines 3--10) until termination criteria are met (line 2).

When computational time of the online optimization is limited, the Newton-type iteration (lines 3--10) may be executed only once per sampling time, as in the real-time iteration scheme \cite{diehl2005real}.
We also empirically found that fixing the barrier parameter of the interior point method and penalty parameters of the soft constraints such as \cite{feller2016relaxed} can enable warmstart of the solution and improve the computational speed, albeit potentially at a slight cost to control performance.
Throughout the following simulation and hardware experiments, the Newton-type iteration was performed only once per sampling time and the barrier and penalty parameters of hard and soft constraints, respectively, were fixed.

\section{EXPERIMENTS}\label{sec:experiments}
\subsection{Full-Centroidal Formulations}
To evaluate the effectiveness of the proposed stochastic/robust NMPC, we conducted simulation and hardware experiments on two distinct robotic platforms: the six-telescopic-wheeled-legged robot Tachyon 3\footnote[1]{Website: \url{https://www.sony.com/en/SonyInfo/research/technologies/new_mobility}}\footnote[2]{Video: \url{https://youtu.be/sorw7o73ydc}} \cite{takasugi2023realtime} and the miniature humanoid robot EVAL-03\footnote[3]{Sony Interactive Entertainment, ``EVAL-03'', Press: \url{https://www.sony.com/en/SonyInfo/News/Press/202210/22-037E/}} \cite{michael2021learning}.
Both platforms are depicted in Fig. \ref{fig:terrainHeightEstimationError}.
Within the stochastic/robust NMPC framework, we employed the full-centroidal NMPC formulation \cite{sleiman2021unified, grandia2023perceptive, katayama2023versatile}, which offers a practical trade-off between model fidelity and computational tractability.
Let $n_J$ be the number of joints of the robot. 
The state $x$ and control input $u$ are defined as $x := \begin{bmatrix} q^{\rm T} & h^{\rm T} \end{bmatrix} ^{\rm T}$ and $u := \begin{bmatrix} \dot{q}_J ^{\rm T} & f^{\rm T} \end{bmatrix} ^{\rm T}$, respectively, where $q \in \mathbb{R}^{6 + n_J}$ is the configuration, $h \in \mathbb{R}^6$ is the normalized centroidal momentum, $\dot{q}_J \in \mathbb{R}^{n_J}$ is the joint velocities, and $f$ denotes the stack of the contact forces.

Since $\dot{q}_J$ is included in the control input, state jumps (i.e., impulsive changes of generalized velocity) are not assumed, i.e., the reset map and its derivative are given as:
\begin{equation}
    x(t_{(m, m+1)} ^+) = x(t_{(m, m+1)} ^-), \quad \nabla_x R_{(m, m+1)} = I_{n_x \times n_x}.
\end{equation}
Note that, although the reset map is the identity, the saltation matrix (\ref{eq:saltation}) is not the identity, and the guard saltation matrix (\ref{eq:guardSaltation}) is not zero, due to the different values of $f_m ^-$ and $f_{m+1} ^+$.
The guard condition consists of the distances between the swing feet and their targeted contact surfaces.
Given a configuration $q$, the guard condition and its derivatives can be computed using forward kinematics and its Jacobian.

For the covariance propagation equation (\ref{eq:covarianceEquation}), we assume:
\begin{equation}\label{eq:centroidalCovariance}
    \Gamma_i := 
    \begin{bmatrix}
        O_{(6 + n_J) \times 6} \\
        I_{6 \times 6} 
    \end{bmatrix},
    \quad
    W_i := \sigma I_{6 \times 6}.
\end{equation}
The parameter $\sigma$ is set to $1.0 \times 10^{-6}$ for Tachyon 3 and $1.0 \times 10^{-8}$ for EVAL-03, to reflect the disparate scales and dynamics of the robots.

The constraints and cost function design for Tachyon 3's NMPC adhere to our preceding work \cite{katayama2023versatile}: we enforced joint position, velocity, and torque limits, friction cone, and foot placement constraints as inequality constraints.
Furthermore, we incorporated collision avoidance constraints between the feet of Tachyon 3 and the environment, which was modeled by a finite number of cuboids.

The NMPC formulation for EVAL-03 largely follows \cite{sleiman2021unified, grandia2023perceptive} while a surface contact of a foot is modeled by four point contacts at the vertices.
We imposed joint position, friction cone, and self-collision avoidance constraints between the left and right legs as inequality constraints.

\subsection{Comparison in Covariance Prediction}\label{subsec:covariancePrediction}
Before conducting simulation and hardware experiments, we assessed how the proposed covariance propagation (\ref{eq:multipleJumpCovarianceEquation})--(\ref{eq:outputCovarianceEquation_P}) predicts future state covariance under contact uncertainties within the stochastic/robust NMPC framework.
We compared the following three methods for covariance propagation with the above NMPC settings of Tachyon 3:
\begin{enumerate}[(a)]
    \item The proposed a priori and a posteriori covariance update (\ref{eq:multipleJumpCovarianceEquation})--(\ref{eq:outputCovarianceEquation_P}).
    \item The proposed a priori covariance update (\ref{eq:multipleJumpCovarianceEquation}).
    \item Covariance update using only the dynamics, i.e., the reset map, in the form of (\ref{eq:covarianceEquation}).
\end{enumerate}
Specifically, case (c) employs the covariance update between contact events as:
\begin{equation}\label{eq:constantJumpCovariance}
    P_{j+1} = \nabla_x R_{(m, m+1)} P_j \nabla_x R_{(m, m+1)} ^{\rm T} + \Gamma_j W_j \Gamma_j ^{\rm T},
\end{equation}
where $\Gamma_j$ and $W_j$ are in the same form as (\ref{eq:centroidalCovariance}).
We evaluated the state covariance predicted at the terminal time of the horizon (set to 2.5 s) after solving an offline trajectory optimization problem with different parameters for covariance prediction in each algorithm.
In cases (a) and (b), we varied the covariance parameter of the guard conditions, i.e., $C_g$.
In case (c), we varied the covariance parameter $W_j$ in (\ref{eq:constantJumpCovariance}) (i.e., $\sigma$ in (\ref{eq:centroidalCovariance}) for $i = j \in \mathcal{J}$). We investigated three types of trotting\footnote[3]{Trot of Tachyon 3 is defined as a walking pattern alternating [LF, RF, MR] feet contact and [MF, LR, RR] feet contact \cite{katayama2023versatile}.} motions in Tachyon 3: 1) moving forward, 2) moving forward with a curve, and 3) moving forward while ascending a step.

\begin{figure}[tb]
    \centering
    \begin{minipage}{1.0 \linewidth}
        \centering
        \includegraphics[scale=0.7]{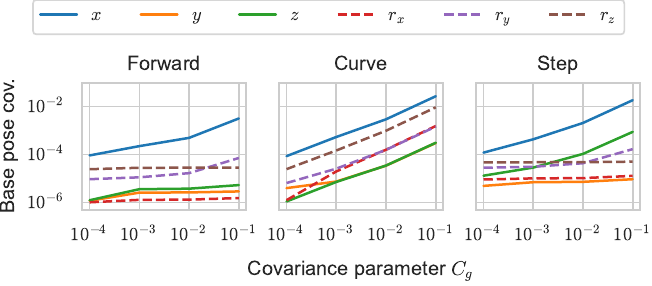}
        \subcaption{A priori and a posteriori covariance updates (\ref{eq:multipleJumpCovarianceEquation})--(\ref{eq:outputCovarianceEquation_P})}
        \vspace{2.5mm}
    \end{minipage}
    \begin{minipage}{1.0 \linewidth}
        \centering
        \includegraphics[scale=0.7]{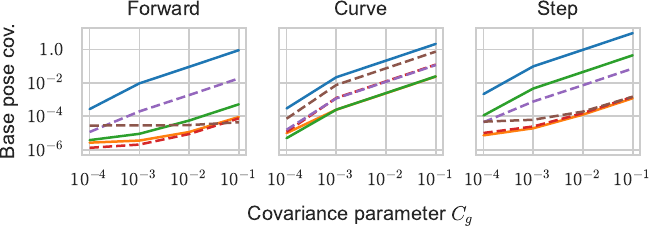}
        \subcaption{A priori covariance update (\ref{eq:multipleJumpCovarianceEquation})}
        \vspace{2.5mm}
    \end{minipage}
    \begin{minipage}{1.0 \linewidth}
        \centering
        \includegraphics[scale=0.7]{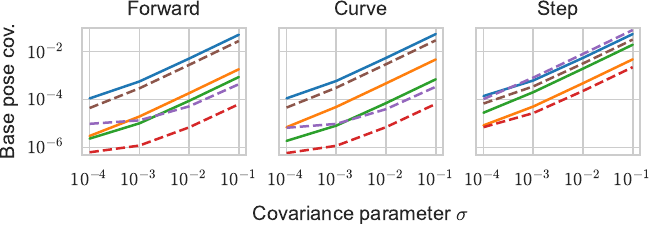}
        \subcaption{Dynamics-based covariance update (\ref{eq:constantJumpCovariance})}
    \end{minipage}
    \caption{Comparison of state covariance propagation methods.
    Plots $x$, $y$, and $z$ denote the covariances of the base position expressed in the world coordinate.
    Plots $r_x$, $r_y$, and $r_z$ represent the covariances of the base rotation around $x$, $y$, and $z$ axes expressed in the base local coordinate, respectively.
    }
  \label{fig:covarianceComparison}
\end{figure}

Fig. \ref{fig:covarianceComparison} illustrates the predicted base pose covariances using the three different methods.  
The state covariance is a $(12 + n_J) \times (12 + n_J)$ matrix, and the plots show the diagonal elements corresponding to the base pose dimensions.  
The comparison between methods (a) and (b) reveals that relying solely on the a priori covariance update (\ref{eq:multipleJumpCovarianceEquation}) leads to an overgrowth of the state covariance.  
This can lead to excessively large back-off terms (\ref{eq:backoff}) and (\ref{eq:backoffJump}), which in turn leads to a conservative solution or makes the numerical optimization challenging to solve.  
Conversely, the results from method (a) demonstrate that the combined use of a priori and a posteriori covariance updates effectively avoids the overgrowth of state covariance.

When contrasting methods (a) and (c), it is evident that the covariance update relying solely on dynamics (\ref{eq:constantJumpCovariance}) leads to a consistent augmentation in covariance across all directions as the covariance parameter $W_i$ is increased.  
On the other hand, the covariance prediction from method (a) is motion-dependent.  
In the case of forward movement, an increase in covariance is predominantly observed in the $x$ direction (the direction of motion), while covariances in other directions, such as $y$ and $z$, remain relatively contained, even as $C_g$ is increased.  
For curvilinear motion, the covariances in directions other than $x$, particularly around the $r_z$ axis (the direction of the curve), are more pronounced than in the forward motion scenario.  
In the scenario of ascending a step, the covariance in the $z$ direction (the direction of ascent) is more significant compared to the forward motion case.  
These results suggest that the proposed method can accurately predict future state covariance in accordance with the anticipated motion (i.e., state and control input trajectories), whereas the dynamics-based covariance update (\ref{eq:constantJumpCovariance}) does not offer such motion-specific predictive capabilities.

\subsection{Simulation Comparison}

To evaluate the efficacy of various MPC methodologies, we executed two sets of simulations under randomized environmental conditions: 1) Tachyon 3 ascending a four-step staircase with height estimation errors, and 2) EVAL-03 traversing a rough terrain area without environment perception.
Both scenarios are depicted in Fig. \ref{fig:terrainHeightEstimationError}.
In both instances, we assessed the success rate of each method across 20 randomized environments.
The stair geometry and height estimation errors (between -4cm and +4cm) were randomized for Tachyon 3, while the rough terrain geometry was randomized for EVAL-03.
The MPC methods compared were:
\begin{itemize}
  \item GS-SMPC (\underline{S}tochastic \underline{MPC} with \underline{G}uard \underline{S}altation Matrix)
  \item SMPC (\underline{S}tochastic \underline{MPC})
  \item HMPC (\underline{MPC} with \underline{h}euristic constraint margins) 
  \item MPC (\underline{MPC} without constraint margins) 
\end{itemize}
SMPC corresponds to the approach presented in a previous study \cite{gazar2023multi}. In GS-SMPC and SMPC, we introduced chance constraints in the form of (\ref{eq:chanceConstraint}) for collision avoidance and joint torque limit constraints in the Tachyon 3 scenario, and for self-collision avoidance and friction cone (specifically the unilateral force) constraints in the EVAL-03 scenario, with satisfaction probabilities set between 0.8 and 0.95.
For GS-SMPC, the covariance of the guard condition was chosen as $C_g = 1.0 \times 10^{-3}$ for Tachyon 3 and $C_g = 1.0 \times 10^{-4}$ for EVAL-03 to encompass contact height uncertainties.
SMPC utilized (\ref{eq:constantJumpCovariance}) for covariance propagation between contact events, with $W_j$ set to values that yielded a state covariance prediction magnitude similar to that of GS-SMPC.
HMPC introduced hand-tuned heuristic margins into the constraints that were treated as chance constraints in GS-SMPC and SMPC.
MPC did not include constant margins.

The horizon lengths were set to 1.5 s and 0.8 s in the Tachyon 3 and EVAL-03 cases, respectively.
The discretization time step was set to 0.02 s in the both cases.
In Tachyon 3 scenario, the controller proposed in our previous work using the control barrier function (CBF) \cite{takasugi2023realtime} was used to track the solution of the MPC while particular CBFs regarding the collision avoidances were turned off to purely evaluate the capabilities of the MPC methods.
In EVAL-03 scenario, a simple inverse-dynamics controller without any equality and inequality constraints was used to compute the joint torques from the MPC solution.

\begin{figure}[tb]
    \centering
    \includegraphics[scale=0.74]{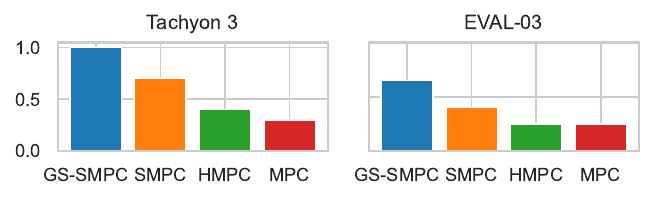}
    \caption{Success rate of four MPC methods for randomized simulations on Tachyon 3 and EVAL-03.}
  \label{fig:successRateComparison}
\end{figure}

\begin{figure}[tb]
    \centering
    \begin{minipage}{1.0 \linewidth}
        \centering
        \includegraphics[scale=0.525]{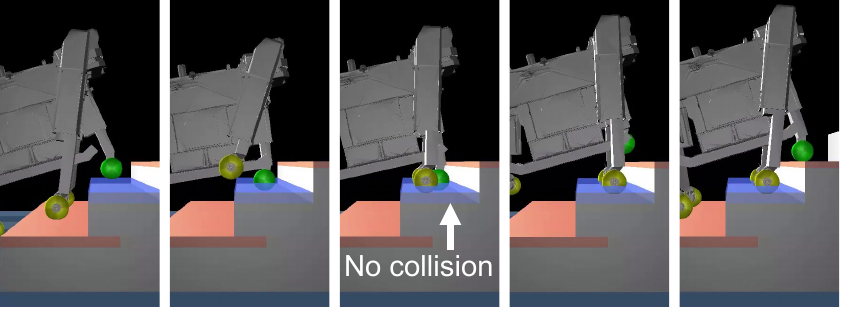}
        \subcaption{GS-SMPC}
        \vspace{1mm}
    \end{minipage}
    \begin{minipage}{1.0 \linewidth}
        \centering
        \includegraphics[scale=0.525]{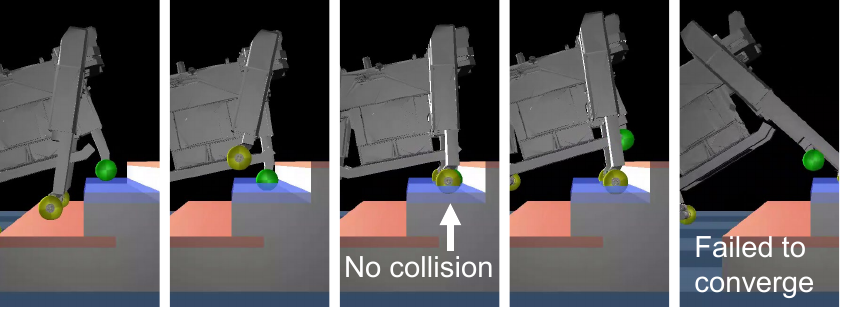}
        \subcaption{SMPC}
        \vspace{1mm}
    \end{minipage}
    \begin{minipage}{1.0 \linewidth}
        \centering
        \includegraphics[scale=0.525]{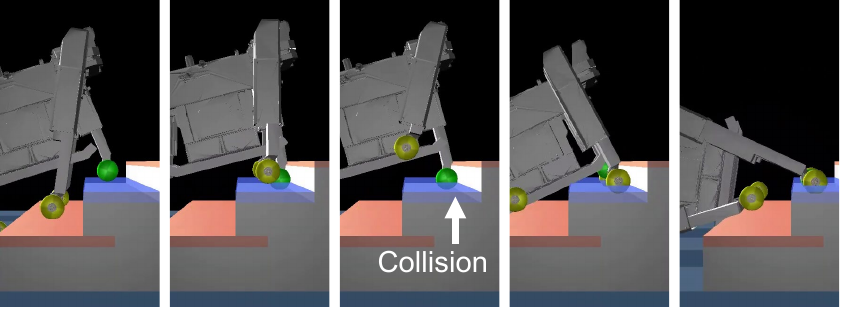}
        \subcaption{HMPC/MPC}
        \vspace{2.0mm}
    \end{minipage}
    \caption{Snapshots of the Tachyon 3 simulations ascending steps under terrain height uncertainties via (a) GS-SMPC, (b) SMPC, and (c) HMPC/MPC.
    The transparent blue boxes indicate the overestimate (e.g., +4 cm overestimate in the second step) and red ones indicate underestimate of terrain heights (e.g., -4 cm underestimate in the first step).
    }
  \label{fig:T3SimSnapshots}
\end{figure}

\begin{figure}[tb]
    \centering
    \includegraphics[scale=0.72]{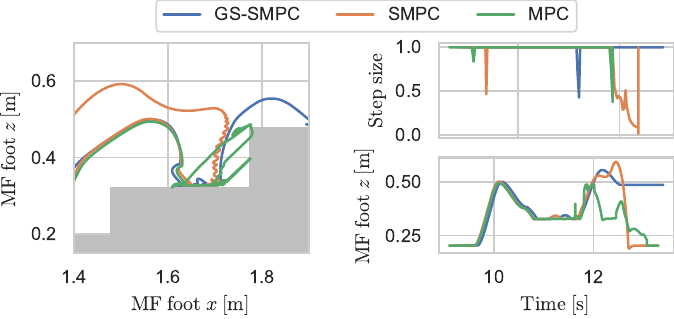}
    \caption{$x$-$z$ trajectories of the MF foot (left) and step sizes of the optimization solver, i.e., convergence behavior (right) during the Tachyon 3 simulations ascending steps.}
  \label{fig:T3SimPlots}
\end{figure}

\begin{figure}[tb]
    \centering
    \begin{minipage}{1.0 \linewidth}
        \centering
        \includegraphics[scale=0.41]{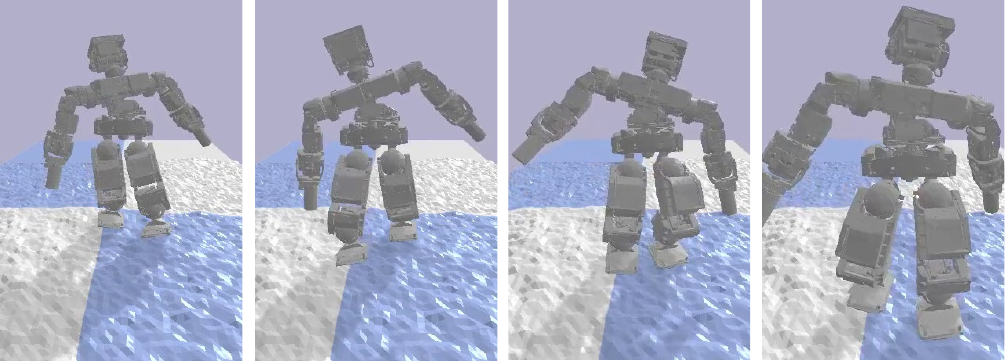}
        \subcaption{GS-SMPC}
        \vspace{1mm}
    \end{minipage}
    \begin{minipage}{1.0 \linewidth}
        \centering
        \includegraphics[scale=0.41]{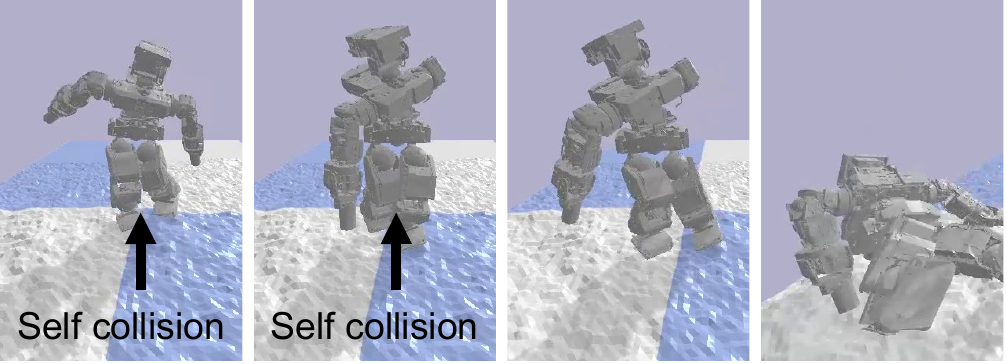}
        \subcaption{MPC}
    \end{minipage}
    \caption{EVAL-03 traversing over rough terrain via GS-SMPC (top row) and MPC (bottom row).
    }
  \label{fig:EvalSimSnapshots}
\end{figure}

Fig. \ref{fig:successRateComparison} shows the success rates of the four methods in the two benchmarks.
The proposed GS-SMPC outperformed the other methods in both scenarios, demonstrating its robustness against contact uncertainties.
Figs. \ref{fig:T3SimSnapshots}--\ref{fig:EvalSimSnapshots} display typical behaviors of Tachyon 3 and EVAL-03 during the experiments.
As illustrated in Figs. \ref{fig:T3SimSnapshots} and \ref{fig:T3SimPlots}, GS-SMPC successfully generated trajectories that avoided collisions while preventing joint torque saturation despite the discrepancies between expected and actual terrain heights.
While SMPC also managed to avoid collisions, it encountered ill-conditioned optimization problems (e.g., too small step size in Fig. \ref{fig:T3SimPlots}) caused by overly large backoff terms, as also illustrated in Fig. \ref{fig:covarianceComparison}.
Is should be noted that even though we attempted SMPC with wide range of $\sigma$ in (\ref{eq:centroidalCovariance}), but could not find successfull one.
HMPC and MPC were unable to accurately predict constraint margins, leading to intense collisions following the terrain height mispredictions.
Fig. \ref{fig:EvalSimSnapshots} also demonstrated effectiveness of GS-SMPC in preventing self-collisions while enhancing the unilateral force constraints under terrain height disturbances.
The robot frequently experienced self-collisions between the left and right legs due to the contact disturbances when using the other MPC methods.
However, as suggested by the success rate for EVAL-03 in Fig. \ref{fig:successRateComparison}, the proposed method still encountered challenges in preventing the robot from falling down, e.g., when a swing foot collided with bumps.

\subsection{Hardware Experiments}

\begin{figure*}[tb]
    \centering
    \begin{minipage}{1.0 \linewidth}
        \centering
        \includegraphics[scale=0.525]{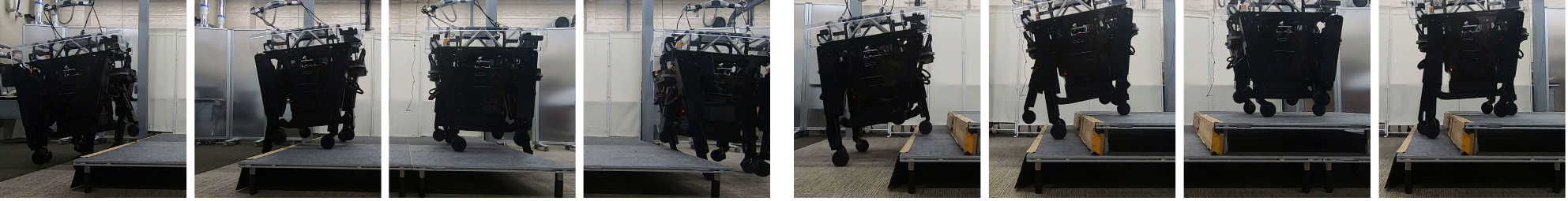}
        \vspace{2mm}
    \end{minipage}
    \begin{minipage}{1.0 \linewidth}
        \centering
        \includegraphics[scale=0.59]{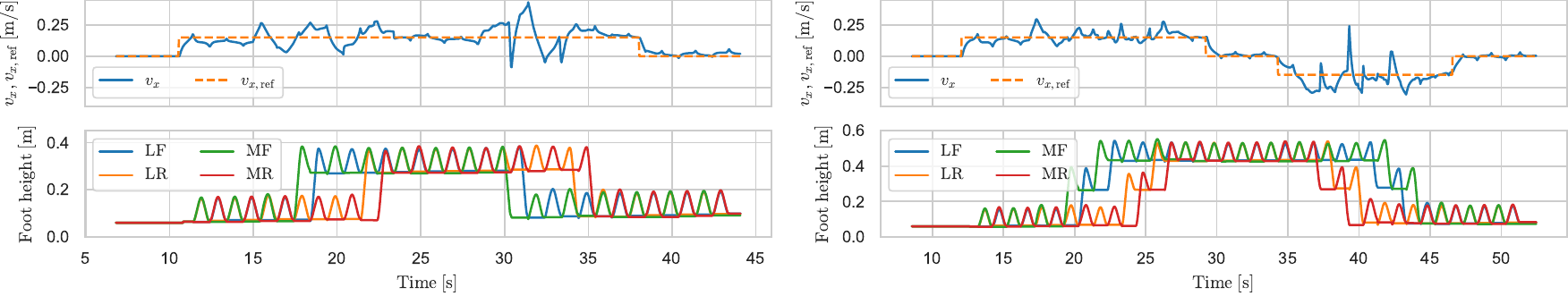}
    \end{minipage}
    \caption{Snapshots and plots of the perceptive locomotion of Tachyon 3 traversing a step (left) and ascending and descending two steps (right) using the proposed stochastic NMPC. $v_x$ denotes the estimated base linear velocity expressed in the local coordinate frame and $v_{x, {\rm ref}}$ does its reference.
    }
  \label{fig:t3HardwareSnapshots}
\end{figure*}

To verify the practical applicability of the proposed method, we conducted hardware experiments on the perceptive locomotion of Tachyon 3.
The software system, including the NMPC implementation, is detailed in \cite{katayama2023versatile}.
The horizon length was set to 1.5 s and the discretization time step was set to 0.02 s.
Although no artificial noise was introduced in the perception as in the simulations mentioned above, the real hardware inherently encompassed noise in the dynamics, contacts, and perception.

Fig. \ref{fig:t3HardwareSnapshots} displays snapshots and plots of Tachyon 3's perceptive locomotion using the proposed GS-SMPC.
Tachyon 3 adeptly navigated over a step and managed to ascend and descend a series of two steps via GS-SMPC, which was implemented on the robot's on-board PC.
The average computational time of GS-SMPC was 29 ms while that of the nominal NMPC (i.e., NMPC without covariance propagation) was 26 ms measured on the on-board PC (CPU: Intel(R) Core(TM) i7-8850 H CPU @ 2.606 GHz).
Thanks to the efficient algorithm of the stochastic NMPC outlined in Section IV, our GS-SMPC showed only a slight increase (approximately 3 ms in this instance) in computational time compared to the nominal NMPC.

\section{CONCLUSIONS}\label{sec:conclusions}
This paper introduced a stochastic/robust NMPC for robust model-based legged locomotion against contact uncertainties.
We leveraged the saltation matrices and output-feedback MPC technique in forecasting the future state covariance, of which the performance had been shown in the numerical studies.
Furthermore, the proposed method could be implemented on a on-board PC by utilizing the efficient zero-order stochastic/robust NMPC algorithm that fully leverages the Riccati recursion to obtain the feedback gains.

On the other hand, as indicated by the simulation study of the humanoid rough terrain walking, the proposed method may not be sufficient in completely preventing the robot from falling down.
To address this limitation, it is crucial to explore other aspects of MPC, such as emergent contact re-planning, from the perspective of contact uncertainty.

\bibliographystyle{IEEEtran}
\bibliography{IEEEabrv, root}

\end{document}